\documentclass[letterpaper]{article} % DO NOT CHANGE THIS
\usepackage{aaai24}  % DO NOT CHANGE THIS
\usepackage{times}  % DO NOT CHANGE THIS
\usepackage{helvet}  % DO NOT CHANGE THIS
\usepackage{courier}  % DO NOT CHANGE THIS
\usepackage[hyphens]{url}  % DO NOT CHANGE THIS
\usepackage{graphicx} % DO NOT CHANGE THIS
\usepackage{natbib}  % DO NOT CHANGE THIS AND DO NOT ADD ANY OPTIONS TO IT
\usepackage{caption} % DO NOT CHANGE THIS AND DO NOT ADD ANY OPTIONS TO IT
\usepackage{algorithm}
\usepackage{algorithmic}
\usepackage{tabularx}
\usepackage{multirow}
\usepackage{appendix}
\usepackage{array}
\usepackage{booktabs}
\usepackage{amsmath,amsfonts}

\usepackage{amsmath} % Include this in your preamble

\usepackage{newfloat}
\usepackage{listings}
\DeclareCaptionStyle{ruled}{labelfont=normalfont,labelsep=colon,strut=off} % DO NOT CHANGE THIS
\floatstyle{ruled}
\newfloat{listing}{tb}{lst}{}
\floatname{listing}{Listing}
\title{Spatial-Related Sensors Matters: 3D Human Motion Reconstruction Assisted with Textual Semantics}
\author{
    Xueyuan Yang \textsuperscript{\rm 1 2},
    Chao Yao\textsuperscript{\rm 1 2}\thanks{Corresponding author.},
    Xiaojuan Ban\textsuperscript{\rm 1 2 3 4 *}
}
\affiliations{
     \textsuperscript{\rm 1}Beijing Advanced Innovation Center for Materials Genome Engineering, Beijing 100083, China.\\
     \textsuperscript{\rm 2}University of Science and Technology Beijing, Beijing 100083, China.\\
    \textsuperscript{\rm 3}Key Laboratory of Intelligent Bionic Unmanned Systems, Ministry of Education, Beijing 100083, China.\\
    \textsuperscript{\rm 4}Institute of Materials Intelligent Technology, Liaoning Academy of Materials, Shenyang 110004, China.\\
   m202210673@xs.ustb.edu.cn,\{yaochao, banxj\}@ustb.edu.cn

}

\usepackage{bibentry}
\begin{document}

\maketitle

\begin{abstract}
Leveraging wearable devices for motion reconstruction has emerged as an economical and viable technique. Certain methodologies employ sparse Inertial Measurement Units (IMUs) on the human body and harness data-driven strategies to model human poses.
However, the reconstruction of motion based solely on sparse IMUs data is inherently fraught with ambiguity, a consequence of numerous identical IMU readings corresponding to different poses. 
In this paper, we explore the spatial importance of multiple sensors, supervised by text that describes specific actions. Specifically, uncertainty is introduced to derive weighted features for each IMU. We also design a Hierarchical Temporal Transformer (HTT) and apply contrastive learning to achieve precise temporal and feature alignment of sensor data with textual semantics.
Experimental results demonstrate our proposed approach achieves significant improvements in multiple metrics compared to existing methods. Notably, with textual supervision, our method not only differentiates between ambiguous actions such as sitting and standing but also produces more precise and natural motion.
%这个abstract不太好，需要改一下
\end{abstract}

\section{Introduction}
%问题定义
Human motion reconstruction is a pivotal technique for accurately capturing 3D human body kinematics, with critical applications in gaming, sports, healthcare, and film production.
One of the prevalent methods in motion reconstruction is the optical-based approach, which involves analyzing images of individuals to ascertain their respective poses \cite{chen2020cross,sengupta2023humaniflow,cao2017realtime}. With the rapid progression of wearable techniques, various sensor devices have also been used to reconstruct human motion.
For example, Xsens \cite{schepers2018xsens} system employs 17 densely positioned IMUs to facilitate the reconstruction of human body poses.
Compared to optical methods, IMUs offer robustness against variable lighting conditions and occlusions, allow for unrestrained movement in both indoor and outdoor environments, and enable the generation of naturalistic human motion.
% Compared to optical sensors, IMUs offer several advantages, including resilience to lighting conditions and occlusions, unrestricted movement for indoor and outdoor activities, and naturalistic motion generation. 
However, the dense placement of wearable IMUs on the body can be intrusive and costly.
\begin{figure}[t]
\centering
\includegraphics[width=1\columnwidth]{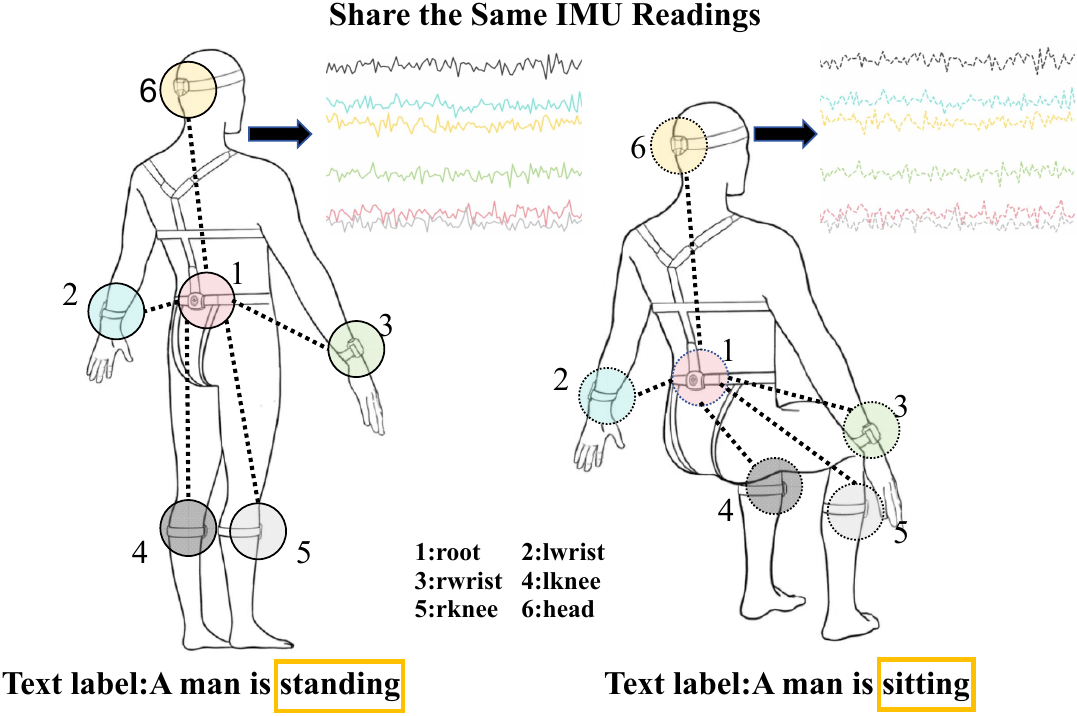} 
\caption{Considering specific postures such as standing and sitting, the rotational data and acceleration output by the sensors are largely invariant. Incorporating additional information such as text can help to address this challenge.}
% The sole deviation recorded is the instantaneous acceleration, which occurs during the transition between these states.}
\label{fig1}
\end{figure}
%加上一些简单的文本可以帮助解决
% Concurrently, the swift advancement of wearable devices has given rise to alternative techniques, which capitalize on various sensor devices for human motion reconstruction.

% Relative to optical sensors, IMUs confer several significant advantages including resilience to lighting conditions and occlusions, the freedom of unrestricted movement for both indoor and outdoor activities, and the generation of more naturalistic motion.
% Furthermore, methodologies grounded in inertial sensors can alleviate data privacy issues commonly associated with optical-based solutions.
% In comparison to optical sensors, IMUs provide several notable benefits. These encompass resistance to lighting conditions and occlusions, unrestrained movement for both indoor and outdoor activities, and the generation of more natural motion.
% Moreover, approaches based on inertial sensors alleviate data privacy concerns linked to optical-based solutions.
%不需要对称，inthis paper  tsa  各个模块  contribution

To address this issue, some methods \cite{huang2018deep,yi2021transpose,jiang2022transformer,von2017sparse,yi2022physical} have deployed sparse IMUs on the body and analyzed temporal signals to model human body poses.
These approaches not only reduce the number and cost of IMUs but also enhance the wearability and minimize invasiveness.
% This methodology offers significant advantages in terms of reducing the number and cost of IMUs, as well as providing enhanced ease of
% wearability and reduced invasiveness.
% Nevertheless, dense IMUs placement on the body remains invasive and costly.no
% Nevertheless, certain issues still limit the application of sparse sensors. Specifically, The existing methods face challenges in differentiating between actions that share similar IMUs output patterns. 
Nevertheless, it should be noted that there are still some limitations that restrict the utilization of sparse sensors. 
% existing methods face challenges in differentiating between actions that share similar IMU output patterns.
% as illustrated in Fig. \ref{fig1}, the standing and sitting actions share the same IMU output patterns, which presents a challenge in accurately differentiating between them.
% 
% Specifically, 
% as depicted in Fig. 1, sparse sensors positioned on the body yield similar rotation matrices and acceleration outputs between standing and sitting actions, which is difficult to accurately differentiate them.
Specifically, motion reconstruction using sparse inertial sensors constitutes an under-constrained problem: distinct postures can yield identical sensor outputs. As illustrated in Fig. \ref{fig1}, the sensors generate similar rotation matrices and acceleration outputs when the subject is sitting and standing, making accurate differentiation between these postures challenging.
Besides, the inherent distinction of spatial relations between IMUs, has rarely been used in previous methods, thereby revealing opportunities for potential enhancements.

% In this paper, inspired by recent advancements in natural language processing and multimodal research, we propose a sensor-based 3D human motion reconstruction with textual supervised framework. 
%明确sensor 我们设计了一个基于对比学习机制  %明确  不能说多模态

% In this paper, we introduce a novel framework for sensor-based 3D human motion reconstruction, leveraging spatial relationships and textual supervision to accurately generate naturalistic human body poses.
In this paper, we introduce a novel framework for sensor-based 3D human motion reconstruction, leveraging spatial relationships and textual supervision to accurately generate naturalistic human body poses.
Sparse sensors are designed to capture the motion characteristics of different body parts. Considering that the correlations among these features contain a crucial priori knowledge about the human body's skeletal structure, our method employs intra-frame spatial attention to model the correlation between IMUs, allowing the model to concentrate on the distinct characteristics of different body regions at one point in time. Moreover, in response to the inherent potential instability of IMU readings, the concept of sensor uncertainty is introduced. This allows for the optimization of sensor outputs and the adaptive adjustment of each sensor's relative contribution. However, relying solely on sensor data is insufficient for resolving the problems of ambiguity. Text, with its rich motion information, can aid the model in identifying human motion states and resolving issues of ambiguity.
Finally, to facilitate better modality fusion, we propose unique modules to align sensor features with text features in both temporal and semantic dimensions.

In the realm of sensor data and text fusion, IMU2CLIP \cite{moon2022imu2clip} bears resemblance to our work, 
aligning images and IMU sensor data with corresponding text using the CLIP \cite{radford2021learning}.
% proposing the alignment of images and IMU sensor readings from smart glasses with their corresponding text within the space of CLIP \cite{radford2021learning}. 
The methodologies diverge in several key respects. IMU2CLIP is designed for modality transitivity, facilitating text-based IMU retrieval, IMU-based video retrieval, and natural language reasoning tasks with motion data. In contrast, our approach underscores the synergistic potential of multimodal information, using text to resolve ambiguities inherent in sparse sensor data. 
Furthermore, to achieve enhanced modality fusion, the Hierarchical Temporal Transformer module was designed, and contrastive learning was employed to ensure temporal and semantic synchronization between the textual and sensor data. Cross-attention mechanisms were then utilized to merge features from both modalities.
Experimental results show that our proposed framework achieves state-of-the-art performance compared with some classical methods, both in quantitative and qualitative measurements.
In summary, our work makes the following contributions:
\begin{itemize}
\item We present a sensor-based approach to 3D human motion reconstruction that is augmented with textual supervision. This method leverages the rich semantic information contained within the text to enhance the naturalness and precision of the modeled human poses.
\item 
We introduce a spatial-relation representation model which computes the correlations between sensors within a frame while also taking into account the uncertainty of each IMU.
% We present a spatial-relation representation model based on IMUs' signals. Specifically, it constructs the spatial relations between the signals, taking into account the uncertainty of each IMU.
% and augment the system's robustness against ambiguous action.
\item 
We design a Hierarchical Temporal Transformer module to achieve temporal alignment between sensor features and textual semantics. A contrastive learning mechanism is also adopted to optimize the alignment between the two modalities in high-dimensional space.
\end{itemize}

\section{Related Work}
\begin{figure*}[!t]
\renewcommand{\baselinestretch}{1.0}
\centering 
\includegraphics[width=7 in, height=3 in]{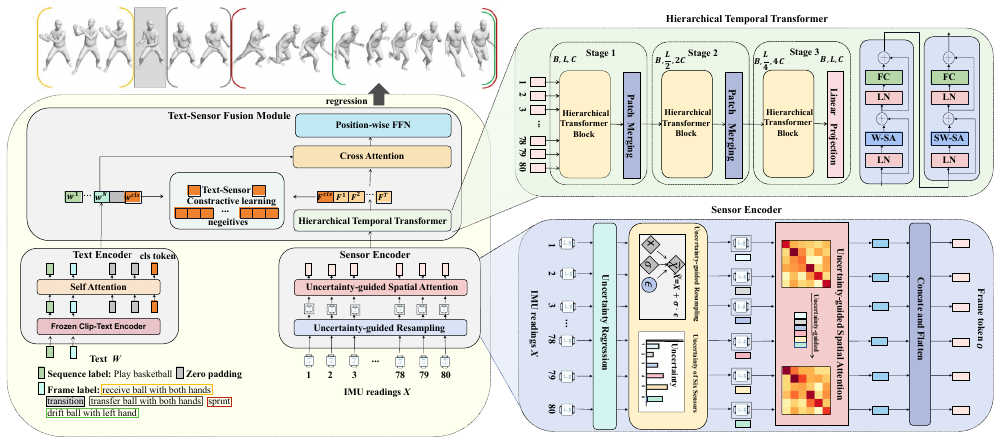}
\caption{Overview of our method. Our model encapsulates three distinct encoders: a Text Encoder, a Sensor Encoder, and a Text-Sensor Fusion Module. The details of the Sensor Encoder and the Hierarchical Temporal Transformer module are illustrated on the right. The schematic of the model output is adapted from \cite{BABEL:CVPR:2021}.}\label{fig2}
\end{figure*}
\subsection{Sensor-based Human Motion Reconstruction}
Full-body sensor-based motion reconstruction is a widely utilized technique in commercial motion capture systems. A prominent example is the Xsens system \cite{schepers2018xsens}, which achieves detailed reconstruction of human movements by equipping the body with 17 strategically placed IMUs.
However, this method presents drawbacks, primarily its invasive impact on human movement due to the intensive IMUs placement, as well as its substantial cost.

Efforts have been made to implement motion reconstruction using sparse IMUs, thereby enhancing the usability of inertial sensor-based motion reconstruction, albeit at the expense of some degree of accuracy. For instance, studies \cite{slyper2008action,tautges2011motion} have achieved human motion reconstruction with as few as four to five accelerometers, by retrieving pre-recorded postures with analogous accelerations from motion reconstruction databases. \cite{von2017sparse} developed an offline system that operates with only six IMUs, optimizing the parameters of the SMPL body model \cite{loper2015smpl} to fit sparse sensor inputs. With the advent of the deep learning era, \cite{huang2018deep} synthesized inertial data from an extensive human motion dataset to train a deep neural network model based on a Bidirectional Recurrent Neural Network that directly mapped IMU inputs to body postures. \cite{yi2021transpose} decomposed body posture estimation into a multi-stage task to improve the accuracy of posture regression through the use of joint locations as an intermediate representation. Moreover, recent methodologies such as \cite{dittadi2021full} and AvatarPoser \cite{jiang2022avatarposer} estimated full-body posture using only head and hand sensors, yielding promising results.

However, reconstructing human motion from a set of sparse IMUs presents an under-constrained problem, where similar sensor readings may correspond to different postures. Some approaches have sought to address this issue to a certain extent through unique network designs. For instance, Physical Inertial Poser \cite{yi2022physical} approximated the under-constrained problem as a binary classification task between standing and sitting, proposing a novel RNN initialization strategy to replace zero initialization. It then distinguished between standing and sitting based on instantaneous acceleration. Transformer Inertial Poser \cite{jiang2022transformer} introduced past history outputs as inputs to differentiate ambiguous actions. 
Other methodologies have explored the integration of multimodal information to impose additional constraints on the model, enhancing the generation of precise poses. For instance, studies \cite{von2018recovering,malleson2017real,von2016human} have significantly improved estimation accuracy by combining inertial sensors with video data, although challenges such as occlusion, lighting issues, and mobility restrictions still persist. Fusion Poser \cite{kim2022fusion} incorporates head height information from a head tracker into the model's input.

% For instance, PIP \cite{yi2022physical} reframes the under-constrained problem as a binary classification task between standing and sitting. It proposes a novel RNN initialization strategy to replace zero initialization and discriminates between standing and sitting based on instantaneous acceleration. TIP \cite{jiang2022transformer} incorporates past output history as input to differentiate ambiguous actions.
% Other methodologies aim to integrate multimodal information to impose additional constraints on the model for accurate posture generation. For example, studies \cite{von2018recovering,malleson2017rea,von2016human} have significantly improved estimation accuracy by combining inertial sensor data with video data, despite persistent challenges such as occlusion, lighting issues, privacy concerns, and mobility restrictions. Fusion Poser \cite{kim2022fusion} partially addresses the problem by combining IMU data with the head height information returned from a head tracker.
\subsection{Textual Semantics in Human Motion Field}
In the burgeoning field of multimodal processing, text, with its rich semantic information and ease of annotation, is increasingly utilized in the human motion domain. Studies such as \cite{guo2022generating,zhang2022motiondiffuse,tevet2022motionclip} can generate high-quality 3D human motions from textual descriptions. 
These findings affirm that texts encapsulate rich motion information. We posit that text supervision could disambiguate actions, thereby enhancing the naturalness and precision of generated motions.

\section{Method}
%逻辑混乱 先讲文本的，再讲运动的
% Our goal is to obtain an accurate human pose from 6 IMUs placed on the user’s legs, wrists, head, and pelvis, and the corresponding semantic information.In adherence to previous research methodologies, we employ the tri-axial acceleration and rotation matrix data outputs derived from each frame of the six sensors as model input.
Our primary target is to reconstruct accurate human poses using data from 6 IMUs placed on the legs, wrists, head, and pelvis (root), coupled with textual supervision. The sensors provide inputs in the form of tri-axial acceleration, $a \in \mathbb{R}^{3}$, and rotation matrices, $R \in \mathbb{R}^{3\times3}$.
%sensor text输入
% To create the text encoder, We have incorporated the first four frozen layers from the CLIP's ViT/B32 model, supplemented by an additional two Transformer layers. 
% The UGRT and UGST are employed to reconstruction spatial correlations among the six sensors and augment the system's robustness, while the Hierarchical Temporal Transformer is utilized to extract temporal dependencies within the motion sequence efficiently.
%解释
% where mcls represents the embedding of the [CLS] token. 
% In the realm of modality fusion, we employ a four-layer transformer predicated upon local window attention as the underpinning for our multimodal encoder. 
% which is not detailed in Figure \ref{fig2}.明确输出 提对比学习
As illustrated in Fig. \ref{fig2}, our framework consists of a Text Encoder, a Sensor Encoder, and a Text-Sensor Fusion module. 
The Text Encoder converts the input text $W$ such as [“receive ball with both hands”, ..., “transition”] into a sequence of embeddings: $\{W^{cls}, W^{1}, ..., W^{N}\}$, where $W^{cls}$ represents the embedding of the [CLS] token, and $N$ denotes the number of text labels. 
For the Sensor Encoder, 
a motion sequence composed of sensor data frames 
$X^{t}=[(a_{root}^t,R_{root}^t),..., (a_{head}^t,R_{head}^t)], t \in [1,T]$ 
is encoded into a sequence of embeddings that contain intra-frame spatial relations: $\{O^{1}, ..., O^{T} \}$. 
% (a_{root}^t,R_{root}^t),...,(a_{head}^t,R_{head}^t) , t \in [1,T]
Within the Text-Sensor Fusion module, these spatial embeddings are then processed by the Hierarchical Temporal Transformer to extract a unified spatio-temporal fusion representation $\{{F^{cls}, F^{1}, ..., F^{T}}\}$, where $F^{cls}$ denotes the embedding of the [CLS] token.
Before applying cross-attention in the fusion process, Text-Sensor contrastive learning is strategically implemented to refine the alignment between the unimodal representations of the two modalities.
Finally, a simple regression head is employed to derive human pose rotational data $q \in \mathbb{R}^{j\times6}$ for $j$ key points (with each rotation encoded by a 6D vector \cite{zhou2019continuity}), the corresponding three-dimensional position $p \in \mathbb{R}^{j\times3}$, and the root's speed data $s \in \mathbb{R}^{3}$.
%这里得改
% We employ contrastive learning to align the features from both sensors and text within a high-dimensional space. These features are then harmoniously fused through cross attention within the Text-Sensor Fusion Module. Subsequently,

\subsection{Text Encoder} 
% We employ the first 4 layers of the frozen CLIP VIT/B32 text encoder in conjunction with two transformer layers. 
% 改 中式英文
% This forms our text encoder, transforming the input text and an additional [CLS] token into a sequence of text embeddings $\{wcls, w^{1}, ..., w^{N}\}$. The $wcls$ token is used to summarize the input text.
We utilize the first 4 layers of the frozen CLIP \cite{radford2021learning} VIT/B32 text encoder, augmented with two additional transformer layers, to form our Text Encoder. Specifically, given a text label sequence $W$, it is initially tokenized and mapped into a sequence of tokens $\widetilde{W}$ using CLIP, with a zero-initialized tensor prepended as the [CLS] token. It is important to note that $W$ provides two kinds of semantic labels: sequence-level and frame-level labels, as defined in the dataset configuration. For frame-level labels, despite each frame having its own text description, they are largely repetitive. For example, the label ``walk'' might apply continuously over a series of frames. To mitigate computational load, only non-repetitive frame-level texts are chronologically ordered as inputs. For sequence labels, if the total number is less than the threshold $M$, we use all sequence labels as input. Otherwise, one-third of the labels are selected based on their temporal information, specifically choosing those that best match the sensor subsequence.
% we select sequence labels equivalent to one-third of the total sequence number that best match the motion subsequence by using temporal information. 
% To differentiate sequence-level and frame-level labels, two learnable group encodings $G$ for each are developed. Additionally, Sinusoidal Position Encoding\cite{vaswani2017attention} $T$ are utilized based on their temporal information. 
To differentiate between sequence-level and frame-level labels, two learnable group position embeddings \(G\) are developed for each. Additionally, Sinusoidal Position Embeddings \cite{vaswani2017attention} 
\(P\) are utilized, with time information computed independently for both the sequence and frame levels, accommodating their unique characteristics.
\begin{equation}
     \overline{W}^{i} = \widetilde{W}^i+P^i +G^{i}, ~~ for~~ i\in [1,N]
\end{equation}
Then the processed features $\overline{W}$ and the [CLS] token are fed into the self-attention layers to better extract textual semantics.
%最后学出了什么textual semantic
\subsection{Sensor Encoder} 
The Sensor Encoder captures the intricate relations within sparse sensors via spatial modeling. It includes a resampling strategy and a spatial attention mechanism, both guided by estimated uncertainty for each IMU.   

\textbf{Uncertainty Estimation:}
First, we estimate the uncertainty for each IMU reading, where the original IMU readings, denoted by $X^{t} \in \mathbb{R}^{6\times(3+3\times3)}$ are fed into an uncertainty regression head, yielding uncertainty $\sigma^{t} \in \mathbb{R}^{72}$ for each channel.

\textbf{Uncertainty-guided Resampling:}
Rather than directly using the original readings $X^{t}$, we resample IMU readings denoted as $\widetilde{X}^{t}$ from a Gaussian distribution $\mathcal{N}(X^{t}, \sigma^{t})$, with $X^{t}$ as the mean and predicted uncertainty $\sigma^{t}$ as the variance. This resampling method ensures that the values with low uncertainty remain largely unchanged, while the values with high uncertainty are resampled, thereby optimizing the sensor data.
Notably, the resampling procedure is only employed during the training. During inference, the uncertainty is simply regressed for each channel, and the original sensor readings $X^{t}$ are utilized as $\widetilde{X}^{t}$.
% The sampling procedure is exclusively employed during the training phase, and during inference, we only regress the uncertainty for each sensor channel and use the original sensor readings as $\widetilde{Y}$.
We apply the reparameterization trick \cite{kingma2022autoencoding} for efficient gradient descent by sampling $\epsilon \sim\mathcal{N}(\textbf{0},\textbf{1})$ to compute $\widetilde{X}^{t}$ as follows: $\widetilde{X}^{t} =X^{t}+\sigma^{t} \cdot \epsilon$ .

\textbf{Uncertainty-guided Spatial Attention (UGSA):}
% Upon acquisition of the i-th frame's sampled IMU readings, denoted as $\widetilde{Y}^{i}$, and their corresponding uncertainties$\sigma^{i}$. We project $\widetilde{Y}^{i}$ into feature embeddings, denoted as ${Z}^{i} \in R^{6\times C}$,where C is the spatial embedding dimension, and equip ${Z}^{i}$ with positional embeddings. Subsequently, these results are forwarded to the Transformer layer within the UGST. 
% We then enrich ${Z}^{i}$ with positional embeddings, and these augmented representations are subsequently forwarded to the Transformer layer embedded within the UGST.
% The layer bears resemblance to the standard Transformer\cite{vaswani2017attention},
% Upon obtaining the sampled IMU readings for the $t$-th frame, represented as $\widetilde{X}^{t}$, along with their corresponding uncertainties $\sigma^{t}$, we project $\widetilde{X}^{t}$ into a feature embeddings space, symbolized as ${Z}^{t} \in \mathbb{R}^{6\times c}$. Here, 6 corresponds to the six sensors, and $c$ signifies the spatial embedding dimension. We then conduct self attention \cite{vaswani2017attention} on ${Z}^{t}$.
% It is noted that for the computation of $t$-th frame's attention between two sensors, denoted as $j$ and $k$, the uncertainty $\sigma_{k}^{t}\in \mathbb{R}^{12}$ (summed over its 12 channels) is taken into account by dividing the attention score by it. 
After sampling the IMU readings for the \(t\)-th frame \(\widetilde{X}^{t}\) with corresponding uncertainty \(\sigma^{t}\), we map \(\widetilde{X}^{t}\) to a \(6 \times c\) feature embedding \(Z^{t}\), where 6 represents the number of sensors and \(c\) signifies the dimension of spatial features. We then conduct self-attention \cite{vaswani2017attention} on ${Z}^{t}$.
It is noted that for the computation of $t$-th frame's attention between two sensors, denoted as $j$ and $k$, the uncertainty $\sigma_{k}^{t}\in \mathbb{R}^{12}$ (summed over its 12 channels) of sensor $k$ is taken into account by dividing the attention score by it. 
\begin{equation}
A_{j,k}^{t}=\frac{\left(Z_j^{t} P^Q\right)\left(Z_k^{t} P^K\right)^T}{\sqrt{c} \cdot \sum{\sigma_k^{t}}}
\end{equation}
where $P^Q,P^K \in \mathbb{R}^{c\times c}$ are the Query and Key projection matrices. This unique alteration ensures that sensors with high uncertainty contribute less when computing spatial correlations.
% For the UGSA module, the output of its $t$-th frame maintains dimensional congruence with the input ${Z}^{t}$.denoted as ${O}^{t} \in \mathbb{R}^{6\times c}$.We then flatten ${O}^{t}$ as ${O}^{t} \in \mathbb{R}^{1\times (6\cdot c=C)}$and concatenate these vectors \{${O}^{1}$, ${O}^{2}$, . . . , ${O}^{T}$ \} from the $T$ input frames as ${O} \in \mathbb{R}^{T\times C}$.
The output of the UGSA module for the \(t\)-th frame, \({O}^{t}\), matches the input dimensions \({Z}^{t} \in \mathbb{R}^{6\times c}\). After flattening \({O}^{t}\) to \(\mathbb{R}^{1\times (C=6c)}\), we concatenate the output vectors from \(T\) frames to form \({O} \in \mathbb{R}^{T\times C}\).
\begin{figure}[t]
\centering
\includegraphics[width=\columnwidth]{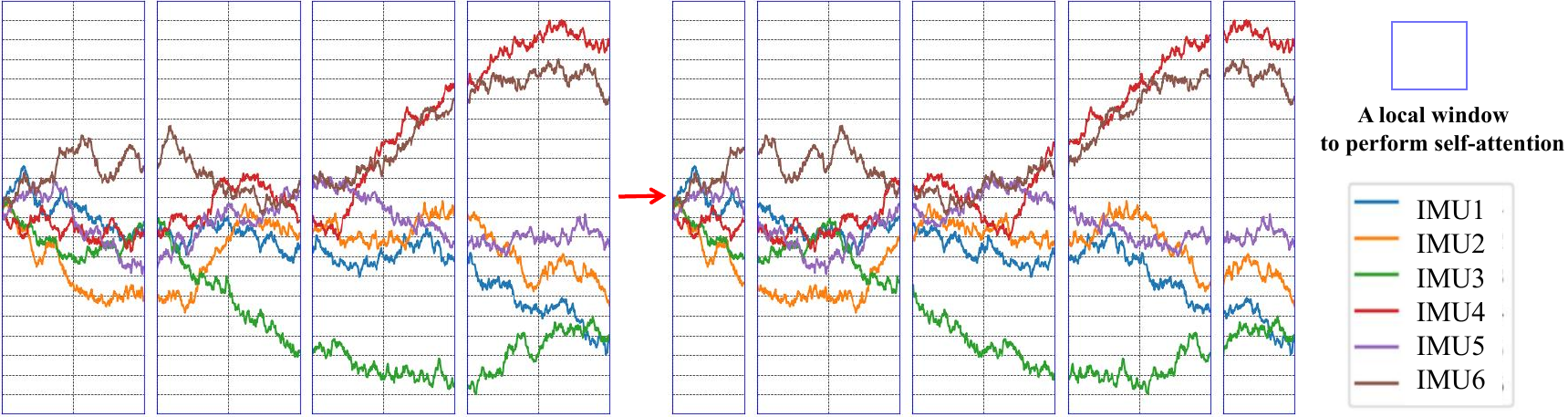} 
% An illustration of the W-SA and the SW-SA. On the left, We divide the sensor sequence into non-overlapping sub-segments based on a predefined window size. Computations for local window attention are then performed within these segments. In a quest to foster interrelations among non-overlapping segments, we introduce a sliding window operation. This operation facilitates a novel window partitioning scheme where the self-attention computation is carried out, as illustrated on the right.
% On the left, W-SA divides the sensor sequence into non-overlapping segments for localized attention. In a quest to foster interrelations among non-overlapping segments, SW-SA implements a sliding window technique, enabling a new partitioning method that enhances self-attention across segments as demonstrated on the right.
\caption{An illustration of window self-attention (left) and shifted window self-attention (right).}
\label{fig3}
\end{figure}

\subsection{Text-Sensor Fusion Module} 
%先去看一下后面的模型设计
The Text-Sensor Fusion Module aligns and fuses bimodal features.
Specifically, we employ a Hierarchical Temporal Transformer to acquire spatiotemporally fused sensor features for temporal synchronization with text features. Subsequently, contrastive learning is used to align the multimodal features in a high-dimensional space, followed by the application of cross-attention for feature fusion.
% The Text-Sensor Fusion Module is designed to bridge the temporal alignment and fusion between the spatial-relation features of IMUs and the textual semantics. We design a Hierarchical Temporal Transformer to obtain the temporal tokens from the spatial-relation features. In addition, contrastive learning and cross attention are employed to align and fuse the temporal tokens and the textual semantics.

\textbf{Hierarchical Temporal Transformer (HTT):} The HTT module is utilized for temporal alignment between sensor features and textual semantics. We hypothesized that information derived from adjacent frames is pivotal for the estimation of the current frame pose. In response to this hypothesis, window self-attention (W-SA) and shifted window self-attention (SW-SA) mechanisms are incorporated to constrain the scope of attention computation, introducing a convolution-like locality to the process.
Furthermore, to integrate information from distant frames and thus extend the receptive field, a patch merge operation is implemented. This approach facilitates the extraction of sensor features at diverse granularity levels and concurrently reduces the computational complexity of the transformer from a quadratic to a linear relationship with the sequence length.
% HTT module incorporates window self-attention (W-SA), shifted window self-attention (SW-SA), and a patch merging strategy. These components foster a linear relationship between computational complexity and input sequence length, enhancing the efficiency of temporal modeling.

Given a window size of $I$, a sensor sequence of length $L$ is divided into $\frac{L}{I}$ non-overlapping subintervals. Local window attention computations are first performed within these subintervals. To create interconnections between these non-overlapping segments, we adopt a shifted window attention module inspired by \cite{liu2021swin}, enabling a new partitioning method that enhances self-attention across segments. The W-SA and SW-SA always appear alternately, constituting a Hierarchical Transformer Block as shown in the top-right corner of Fig. \ref{fig2}.

% Upon completion of the attention computation, the original sequence order is reinstated via a window reversal operation.
% It's important to note that the local window attention and the shifted window attention always appear in alternating pairs. The associated computation methodology is as follows:
% \begin{align}
% & \hat{\mathbf{s}}^l=\mathrm{W}-\operatorname{SA}\left(\mathrm{LN}\left(\mathbf{s}^{l-1}\right)\right)+\mathbf{s}^{l-1}, \\
% & \mathbf{s}^l=\operatorname{MLP}\left(\mathrm{LN}\left(\hat{\mathbf{s}}^l\right)\right)+\hat{\mathbf{s}}^l, \\
% & \hat{\mathbf{s}}^{l+1}=\operatorname{SW-SA}\left(\mathrm{LN}\left(\mathbf{s}^l\right)\right)+\mathbf{s}^l, \\
% & \mathbf{s}^{l+1}=\operatorname{MLP}\left(\mathrm{LN}\left(\hat{\mathbf{s}}^{l+1}\right)\right)+\hat{\mathbf{s}}^{l+1},
% \end{align}

% In our implementation, $LN$ denotes LayerNorm, and $MLP$ refers to the Multilayer Perceptron. $ \hat{\mathbf{s}}^l$ and $\mathbf{s}^l$ symbolize the output features of the (shifted) window attention module and the MLP module for block $l$, respectively. During attention computations for W-SA and SW-SA, we also introduce relative positional encoding as per the approach outlined in \cite{liu2021swin}.

\begin{figure}[h]
\renewcommand{\baselinestretch}{1.0}
\centering 
\includegraphics[width=0.92\columnwidth]{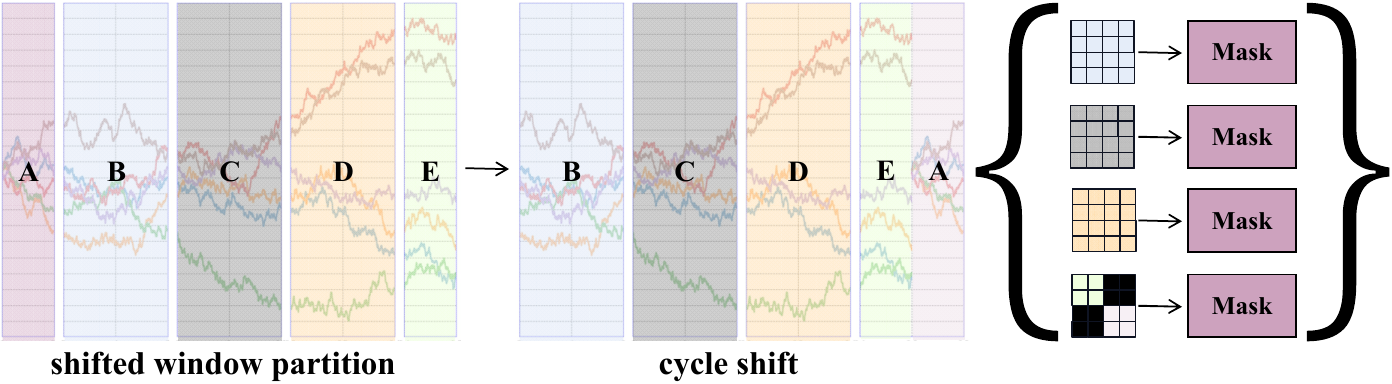}
\caption{An efficient methodology for batch computation of self-attention within the context of shifted window partitioning.}\label{fig4}
\end{figure}
When applying shifted window attention to temporal sequences, the window count increases from $\frac{L}{I}$ to $\frac{L}{I} + 1$, resulting in some windows being smaller than $I$. To address this, we introduce a batch computation with a leftward cyclic shift, depicted in Fig. \ref{fig4}. This shift can produce windows with non-contiguous sub-windows. We tackle this by designing a masking mechanism that restricts self-attention to within each sub-window, maintaining the number of batched windows and ensuring computational efficiency. After the computation, the original sequence order is restored.
% In applying shifted window attention to temporal data, we encounter a challenge: the number of windows changes from $\frac{L}{I}$ to $\frac{L}{I}+1$, leading to some windows with a size less than $I$. To overcome this, we propose an enhanced batch computation approach, characterized by a leftward cyclic shift, as illustrated in Fig. \ref{fig7}. After the shift, a 
% window may encompass non-contiguous sub-window within the feature map. In response to this characteristic, we design a masking mechanism that confines the self-attention computation to each individual sub-window.
 % requiring a masking mechanism to restrict self-attention computation to each individual sub-window.
% This strategy ensures the number of batched windows aligns with the traditional window partitioning, preserving computational efficiency.

In the patch merge operation, each procedure consolidates two adjacent tokens into one, effectively halving the token count and doubling each token's dimensionality. These transformed tokens are then fed into the subsequent stages. Within the final stage, the patch merge is omitted, and tokens are restored to their original count and dimensions through linear projection and reshaping. 
Within a sensor sequence, we map the output features $F \in \mathbb{R}^{T\times C}$ to a feature with dimensions $1\times C$, serving as the [CLS] token. This [CLS] token, in conjunction with $F$, forms the cumulative output $\{F^{cls}, F^{1}, ..., F^{T}\}$, which encompasses the spatio-temporal features.

%基本照搬了swin的说法，后面可能需要改 这里还有两处未说，一个是attention mask的设置，另一个是cls token是如何处理的

%这里要详细叙述一下text encoder的性质，尤其是位置编码部分 组编码，怎么筛选的序列编码  reverse操作还没有说

% Given Motion feature $\{mcls, v^{1}, ..., v^{f} \}$ and Text feature $\{wcls, w^{1}, ..., w^{N}\}$ , we remove the CLS token and use them as inputs to the multimodal fusion module. In order to reduce computational cost, we also employ windowed attention. The motion sequence features are divided into non-overlapping sub-sequences according to window size$W$. When calculating cross-attention, each motion sub-sequence is used as a query, and the concatenation of the motion sub-sequence with the comprehensive semantic features of the entire motion sequence is used as key for the cross-attention computation.

% The output of the fusion encoder consists of rotational data $q_{t} \in R^{j\times6}$ for j key points, their corresponding three-dimensional positional information $p_{t} \in R^{j\times3}$, and the velocity data of the root node $v_{t} \in R^{3}$.
% we omit the [CLS] token and feed them into the Text-Sensor fusion module.

\textbf{Feature Fusion:}
Given the sensor features set $\{{F^{cls}, F^{1}, ..., F^{T}}\}$ and the text features set $\{{W^{cls}, W^{1}, ..., W^{N}}\}$, we apply contrastive learning to align these features in a high-dimensional joint space, utilizing the [CLS] tokens as anchors.
Subsequently, the sensor features are fused with textual features through cross-attention. Corresponding group embeddings and temporal position embeddings are designed for both textual and sensor features.
% To reduce computational demands, we partition the sensor sequence features into distinct sub-sequences for window attention like W-SA. During cross-attention computation, each sensor sub-sequence is used as a query, and its combination with the entire motion sequence serves as the key.
% Subsequently, the features from the sensor encoder, after self-attention computation, are fused with the textual features through cross-attention. Specially, for both the textual and motion features, we have designed corresponding group embeddings and temporal position embeddings\cite{vaswani2017attention} for each. 
% Moreover, to decrease computational demands, We partition the sensor sequence features into distinct sub-sequences to use window attention like W-SA on it. During the cross-attention computation, we use each sensor sub-sequence as a query. The combination of the sensor sub-sequence and the comprehensive semantic features of the entire motion sequence serves as the key.

% The output from the fusion encoder comprises rotational data $q_{t} \in R^{j\times6}$ for j key points, the corresponding three-dimensional position $p_{t} \in R^{j\times3}$, and the root node's velocity data $v_{t} \in R^{3}$.

\subsection{Losses} 
%这里少了不确定度的loss设计
We train our model with three objectives: uncertainty learning on the Sensor Encoder, Text-Sensor contrastive learning on the unimodal encoders and recon loss on the Text-Sensor Fusion module. 
The relevant equations are presented below. The parameters $\delta$, $\gamma$, $\lambda$, $\alpha$, $\beta$ are used to balance the different loss weights.

\textbf{Uncertainty Loss:} We aim to estimate the uncertainty of the input IMU data. Inspired by \cite{kendall2017uncertainties}, we set our uncertainty estimation loss as:
\begin{equation}
\mathcal{L}_\sigma = \frac{\delta}{T} \sum_{t=1}^T(\left\|\frac{({q}^t-\hat q^t)}{ \sum_{j=1}^6\sigma_j^{t}}\right\|^2 + \left\|\frac{({p}^t-\hat p^t)}{ \sum_{j=1}^6\sigma_j^{t}}\right\|^2 +\sum_{j=1}^6\left\|\sigma_j^{t}\right\|^2)
\end{equation}
The term $\sigma_j^{t}$ denotes the uncertainty of the $j$-th sensor at the $t$-th frame. The terms $||{q}^t -\hat q^t||^2$ and $||{p}^t - \hat p^t||^2$ represent the squared discrepancies between the predicted and true values of the joint rotation angles and the joint positions for the $t$-th frame, respectively.

\textbf{Contrastive Loss:} We use text-sensor contrastive learning to learn better unimodal representations before fusion. Given a batch of $B$ text-sensor pairs, the model learns to maximize the similarity between a sensor sequence and its corresponding text while minimizing the similarity with the other $B-1$ texts in the batch, and vice versa. 
% As illustrated by Equation \eqref{e4},\eqref{e5}.
\begin{align}
\mathcal{L}_{\text{contrastive}} = -\frac{\gamma}{2B} \sum_{i=1}^{B} \left(H1 + H2 \right)
\label{e4}
\end{align}
where
\begin{align}
H1 & = \log \frac{e^{s_{i, i} / \tau}}{\sum_{j=1}^{B}e^{s_{i, j} / \tau}}, \
H2 & = \log \frac{e^{s_{i, i} / \tau}}{\sum_{j=1}^{B}e^{s_{j, i} / \tau}}
\label{e5}
\end{align}

The $s_{i, j}$ represents the similarity calculated by cosine similarity between the $i$-th sensor sequence and the $j$-th text, and $\tau$ is a learnable temperature parameter that controls the concentration of the distribution.

\textbf{Recon Loss:} Our model is optimized to encapsulate motion characteristics by minimizing the $L_2$ losses on joint orientations $q$, joint locations $p$, and root speed $s$, as shown in Equations \eqref{e6} and \eqref{e7}.
\begin{equation}
\mathcal{L}_{\text{recon}} = \lambda \cdot {D}(q, \hat{q})+ \beta \cdot {D}(p, \hat{p}) + \alpha \cdot {D}(s, \hat{s}) 
\label{e6}
\end{equation}
where
\begin{equation}
{D}(x, \hat{x}) = \frac{1}{T}\sum_{t=1}^T\left|x^t-\hat{x}^t\right|^2
\label{e7}
\end{equation}
calculates the discrepancy between the model's predicted values $x^t$ and the true values $\hat{x}^t$ for the $t$-th frame.
 
The full objective of our model is:
 \begin{equation}
 \mathcal{L}=\mathcal{L}_\sigma+\mathcal{L}_{\text {contrastive}}+\mathcal{L}_{\text {recon}}
\end{equation}

\section{Experiment}

\begin{table*}[ht]
	\fontsize{10}{12}\selectfont
	\resizebox{\linewidth}{!}{
	\begin{tabular}{c|ccccc|ccccc}
		\hline
		& \multicolumn{5}{c|}{Totalcapture} & \multicolumn{5}{c}{DIP-IMU} \\
		\cline{1-6} \cline{7-11}
		Method& SIP Err ($\rm{deg}$) & Ang Err ($\rm{deg}$) & Pos Err ($\rm{cm}$) & Mesh Err ($\rm{cm}$) & Jitter ($10^2\rm{m}/\rm{s}^3$) &
		SIP Err ($\rm{deg}$) & Ang Err ($\rm{deg}$) & Pos Err ($\rm{cm}$) & Mesh Err ($\rm{cm}$) & Jitter ($10^2\rm{m}/\rm{s}^3$) \\
		\hline
		
		SIP   & -- & -- &   -- & -- & -- & 
		       21.02 ($\pm$9.61) &  8.77 ($\pm$4.38) &  6.66 ($\pm$3.33) &  7.71 ($\pm$3.80) &  3.86 ($\pm$6.32) \\
		DIP  & -- & -- &   -- & -- & -- &
		       16.36 ($\pm$8.60) & 14.41 ($\pm$7.90) &  6.98 ($\pm$3.89) &  8.56 ($\pm$4.65) & 23.37 ($\pm$23.84) \\
		Transpose & 12.30($\pm$5.90) & 11.34 ($\pm$4.84) & 4.85 ($\pm$2.63) & 5.54 ($\pm$2.89) & \textbf{1.31 ($\pm$2.43)} &
		       13.97 ($\pm$6.77) & \textbf{7.62 ($\pm$4.01)} & 4.90 ($\pm$2.75) & 5.83 ($\pm$3.21) & \textbf{1.19 ($\pm$1.76)} \\ \hline
      Ours & \textbf{7.92 ($\pm$4.38)} & \textbf{9.35 ($\pm$4.10)} & \textbf{3.70 ($\pm$2.03)} & \textbf{4.32 ($\pm$2.29)} & 1.74 ($\pm$1.55) &
		       \textbf{13.34 ($\pm$6.71)} &  8.33 ($\pm$4.70) & \textbf{4.71 ($\pm$2.72)} & \textbf{5.75 ($\pm$3.29)} & 1.81 ($\pm$1.72) \\
		\bottomrule
	\end{tabular}}
 \caption{
	    In offline settings, our method is evaluated against SIP, DIP, and Transpose on the Totalcapture and DIP-IMU datasets, focusing on the assessment of body poses.
	    The mean values, along with the standard deviations (enclosed in parentheses), for the sip error, angular error, positional error, mesh error, and jitter error, are presented in the report. Bold numbers indicate the best performing entries.
	}
	\label{table1}
\end{table*}

\begin{table*}[ht]
	\fontsize{10}{12}\selectfont
	\resizebox{\linewidth}{!}{
	\begin{tabular}{c|ccccc|ccccc}
		\hline
		& \multicolumn{5}{c|}{Totalcapture} & \multicolumn{5}{c}{DIP-IMU} \\
		\cline{1-6} \cline{7-11}
		Method& SIP Err ($\rm{deg}$) & Ang Err ($\rm{deg}$) & Pos Err ($\rm{cm}$) & Mesh Err ($\rm{cm}$) & Jitter ($10^2\rm{m}/\rm{s}^3$) &
		SIP Err ($\rm{deg}$) & Ang Err ($\rm{deg}$) & Pos Err ($\rm{cm}$) & Mesh Err ($\rm{cm}$) & Jitter ($10^2\rm{m}/\rm{s}^3$) \\
		\hline
		DIP  & -- & -- & -- & -- & -- &
			   17.10 ($\pm$9.59) & 15.16 ($\pm$8.53) & 7.33 ($\pm$4.23) & 8.96 ($\pm$5.01) & 30.13 ($\pm$28.76) \\
    PIP  & -- & -- & -- & -- & -- &
			   15.02  &8.73 & 5.04 &5.95 & \textbf{2.4} \\
    TIP  & 11.74($\pm$6.75) &  11.57($\pm$5.12) &  5.26($\pm$3.00) & 6.10($\pm$3.44)& 9.69($\pm$6.68) &
			  15.33($\pm$8.44)  & 8.89($\pm$5.04) & 5.22($\pm$3.32) &6.28($\pm$3.89) & 10.84($\pm$6.87) \\
      Transpose& 13.65 ($\pm$7.83) &11.84 ($\pm$ 5.36) & 5.64 ($\pm$ 3.42) &6.35 ($\pm$ 3.70) & \textbf{8.05  ($\pm$11.70)}&  16.68 ($\pm$8.68) &8.85 ($\pm$ 4.82) &5.95 ($\pm$ 3.65) & 7.09 ($\pm$ 4.24) &6.11  ($\pm$7.92) \\ \hline
         	Ours & \textbf{9.67 ($\pm$5.12)} & \textbf{10.49 ($\pm$ 4.55)} & \textbf{4.36 ($\pm$ 2.37)} & \textbf{5.05 ($\pm$ 2.69)} & 13.30 ($\pm$16.86) &
		       \textbf{14.18 ($\pm$7.14)} & \textbf{8.25 ($\pm$ 4.45)} & \textbf{4.76 ($\pm$ 2.76)} & \textbf{5.80 ($\pm$ 3.26)} & 14.41  ($\pm$17.18) \\
		\bottomrule
	\end{tabular}}
 \caption{
	    In online settings, our method is evaluated against DIP, PIP, TIP, and Transpose on the Totalcapture and DIP-IMU datasets, focusing on the assessment of body poses. Bold numbers indicate the best performing entries.
	}
	\label{table2}
\end{table*}

\subsection{Dataset Setting}
% In our experiment, we utilized two types of data: human motion datasets containing IMU data, and text sequences containing semantic annotations corresponding to the motion sequences.
Our experiment employed two types of data: sensor data captured during human motion and the corresponding textual annotations.
% For the semantic information, we employed the Babel dataset\cite{BABEL:CVPR:2021}. Babel is a comprehensive dataset featuring language labels that elucidate the actions performed in motion reconstruction (mocap) sequences. It provides annotations for approximately 43 hours of mocap sequences from AMASS\cite{mahmood2019amass}. The labelling operates on two levels of abstraction: sequence labels, which outline the overall action in the sequence, and frame labels, which detail every action in each sequence frame. Each frame label is meticulously aligned with the duration of the corresponding action in the mocap sequence, enabling the overlap of multiple actions.It is worth noting that since Babel does not provide semantic annotations for the DIP dataset, we manually annotated the DIP dataset at a sequence level. However, these semantic details are less comprehensive than those provided by Babel.

We utilized the Babel dataset \cite{BABEL:CVPR:2021} for semantic annotations, which provides two levels of text labels for around 43 hours of AMASS mocap sequences \cite{mahmood2019amass}: sequence labels describe the overall actions, while frame labels detail each action per frame. For the DIP-IMU dataset \cite{huang2018deep}, which lacks Babel's semantic annotations, we manually added sequence-level labels, albeit less comprehensive.

% Regarding the motion data, the scarcity of real datasets, coupled with the extensive data demands intrinsic to the deep learning training process, led us to adhere to previous methodologies. As a result, we synthesized inertial data for the AMASS dataset, which is significantly larger and contains more variations.We utilized the synthesized dataset in conjunction with the real dataset for training. The detailed dataset configuration is as follows:
Regarding the motion data, given the scarcity of real datasets and the extensive data requirements inherent in deep learning, we followed previous method \cite{jiang2022transformer} and synthesized more diverse inertial data from the extensive AMASS dataset. This enriched synthesized data, combined with real data, was used for training. The configuration details of the motion datasets are as follows:

\textbf{AMASS:} The AMASS dataset unifies various motion reconstruction datasets. We synthesized a subset of AMASS, incorporating the CMU, Eyes Japan, KIT, ACCAD, DFaust 67, HumanEva, MPI Limits, MPI mosh, and SFU datasets.

\textbf{DIP-IMU:} The DIP-IMU dataset comprises IMU readings and pose parameters from approximately 90 minutes of activity by 10 subjects. We reserved Subjects 9 and 10 exclusively for evaluation and utilized the rest for training.

\textbf{Totalcapture:} The Totalcapture dataset \cite{trumble2017total} 
comprises 50 minutes of motion captured from 5 subjects. Following previous works, we used real IMU data for evaluation, but ground truth and synthesized IMU readings were still integrated into the training set. Due to missing semantic annotations from Babel in some sequences, only 27 fully annotated sequences were utilized.
\subsection{Metric}

For a fair comparison, we adhered to the evaluation methodology previously used for \cite{yi2021transpose}. We used five metrics for pose evaluation: 1) SIP error, which measures the average global rotation error of the limbs in degrees; 2) Angular error, the average global rotation error of all body joints, also in degrees; 3) Positional error, the average Euclidean distance error of all joints, with the spine aligned, measured in centimeters; 4) Mesh error, the average Euclidean distance error of the body mesh vertices, with the spine aligned, also in centimeters; 5) Jitter error, the average jerk of all body joints in predicted motion, which reflects motion smoothness.

\subsection{Training Details}

The entire training and evaluation regimen was conducted on a system equipped with 1 Intel(R) Xeon(R) Silver 4110 CPU and 1 NVIDIA GeForce RTX 2080 Ti GPU. Our model was developed using PyTorch 1.13.0, further accelerated by CUDA 11.6. Our model configuration sets the input sequence length $T$ at 80 frames, with a window and shifted size of 20 and 10 frames, respectively, and a threshold of $M$ being 15. The training process, utilizing a batch size of 40, incorporates the Adam optimizer \cite{kingma2014adam} initialized with a learning rate of $2e^{-5}$. To balance the magnitude of the loss, we set $\lambda$ and  $\alpha$ to 1, $\beta$ to 10, $\delta$ to 0.1, and $\gamma$ to 0.01. We focus on regressing information for the 15 major joints as defined in the SMPL model, instead of all joints. Additionally, we apply a moving average with a window size of 15 to the model's output, enhancing the smoothness of the predicted poses.

\subsection{Comparisons}

Quantitative and qualitative comparisons with SIP \cite{von2017sparse}, DIP \cite{huang2018deep}, Transpose \cite{yi2021transpose}, PIP \cite{yi2022physical} and TIP \cite{jiang2022transformer} on the Totalcapture and DIP-IMU datasets. In this comparison, we utilized the best-performing models published by the authors. For TIP, the authors employed a human body format different from ours. Therefore, we converted TIP's output into our format before conducting the comparison. 
% Specifically, as stated earlier, we diverge from previous setups by exclusively using the portion of the Totalcapture dataset that contains semantic annotations for our tests.
\begin{figure}[h]
\centering
\includegraphics[width=\columnwidth]{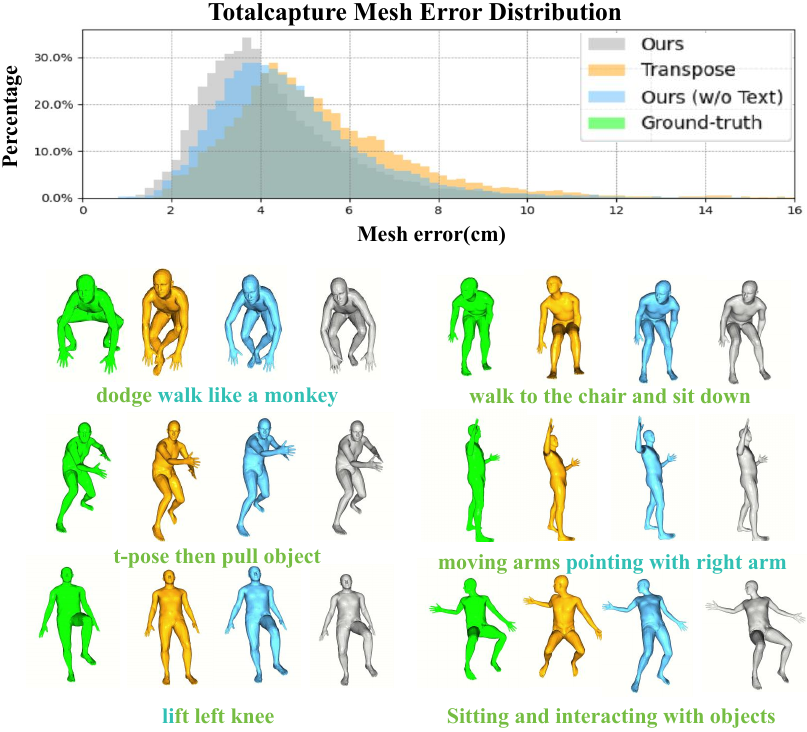} 
% Reduce the figure size so that it is slightly narrower than the column. Don't use precise values for figure width.This setup will avoid overfull boxes.
\caption{Mesh error distribution and qualitative comparisons between our method (with/without text) and Transpose. The text description of the motion is provided below, with the sequence label illustrated in green and the frame label presented in blue.}
\label{fig5}
\end{figure}
The results for the Totalcapture and DIP datasets setting in offline mode are presented in Table \ref{table1}.
Unlike previous methods, our approach does not consider all IMU readings when estimating the current pose. However, our method achieves satisfactory results after integrating semantic information. The performance of our method on the DIP dataset is not as impressive as on the Totalcapture dataset, which can be attributed to the DIP dataset's fewer and less detailed semantic annotations. 
% In Fig 4, we illustrate the comparison of the mesh error distribution and qualitative results between TransPose, which is currently the state-of-the-art (SOTA) method in offline mode, and our proposed method on the Totalcapture dataset. We have selected a few representative visual results to demonstrate the effectiveness of our method. 
As shown in Fig \ref{fig5}, our method excels in processing ambiguous actions like standing and sitting, and is adept at capturing finer details, such as the accurate alignment of hands and feet with the ground truth. This demonstrates a more natural, realistic, and precise performance.

% Our method excels at handling nuanced postures, as shown in Fig 6. Our motion closely aligns with the ground truth, demonstrating more natural, realistic, and precise performance.

It is worth noting that our full model cannot reconstruct human motion in real-time due to the requirement for semantic annotation. Therefore, we employ only the Sensor Encoder and the HTT module for the evaluation in real-time mode. Our method accesses 70 past frames, 5 current frames, and 5 future frames through a sliding window approach, with a tolerable latency of 83 ms. As shown in Table \ref{table2}, despite the absence of semantic information, our method still achieved superiority on multiple metrics, thereby validating the effectiveness of our network design.

The performance of our approach on the jitter metric is not as robust as other metrics, primarily owing to a constrained receptive field from the sliding window mechanism and the patch merging operation, which combines adjacent tokens into a single token. However, we posit that jitter,
unlike the other four pose-accuracy metrics, isn’t as critical. This perspective is based on the observation that visual discrepancies due to jitter are less noticeable when comparing our method with other approaches, while variations in pose precision are notably apparent.

\subsection{Ablation} 
We perform three ablations to validate our key design choices: (1) without text semantic information; (2) without the Uncertainty-guided Spatial Attention (UGSA) module; (3) without the Hierarchical Temporal Transformer (HTT) module. Table \ref{table3} summarizes the results on the Totalcapture dataset (offline).
Ablation experiments underscore the efficacy of our methodological design, with the integration of semantic information being the most salient contribution, followed by the implementation of UGSA and the HTT module.
% \begin{table}[h]
% \fontsize{10}{12}\selectfont
% \centering

% \renewcommand{\arraystretch}{1.7} % 调整行高的倍数
% \resizebox{\columnwidth}{!}{
%  \begin{tabular}{>{\centering\arraybackslash}m{2.5cm}|>{\centering\arraybackslash}m{2.5cm}|>{\centering\arraybackslash}m{2.5cm}|>{\centering\arraybackslash}m{2.5cm}|>{\centering\arraybackslash}m{2.5cm}}
%  \midrule
%  \multicolumn{5}{c}{Totalcapture}\\ % 在Totalcapture周围添加竖线和顶部水平线
%  \hline % 在Totalcapture下添加水平线
% Metric & w/o Text& w/o UGSA & w/o HTT & Ours \\\hline
% SIP Err(deg) &9.21(+/-4.75) &8.67(+/-4.73)  &  8.35(+/-4.57) &\textbf{7.92(+/-4.38)} \\ \hline
% Ang Err(deg) & 10.30(+/-4.43) & 9.94(+/-4.37) & 9.70(+/-4.29)&\textbf{9.35(+/-4.10)} \\ \hline
% Pos Err(cm) & 4.19(+/-2.23) & 4.04(+/-2.22) & 3.89(+/-2.10)&\textbf{3.70(+/-2.03)} \\ \hline
% Mesh Err(cm) &4.86(+/-2.54) &4.67(+/-2.50) &4.52(+/-2.35) &\textbf{ 4.32(+/-2.29)} \\ \hline
% Jitter ($10^{2}m/s^{3}$) & 1.87(+/-1.60) & 1.70(+/-1.55) &\textbf{0.44(+/-1.21)}&1.74(+/-1.55) \\ \hline

% \end{tabular}}
% \caption{Evaluation of Ablation Models on the  Totalcapture Dataset. Bold numbers indicate the best performing entries.}\label{table3}
% \end{table}

\begin{table}[h]
\fontsize{12}{14}\selectfont
\centering
\renewcommand{\arraystretch}{1.7} % Adjust the row height multiplier
\resizebox{\columnwidth}{!}{
 \begin{tabular}{
 >{\centering\arraybackslash}m{2.5cm}|>{\centering\arraybackslash}m{2.5cm}|>{\centering\arraybackslash}m{2.5cm}|>{\centering\arraybackslash}m{2.5cm}|>{\centering\arraybackslash}m{2.5cm}|>{\centering\arraybackslash}m{2.5cm}}
 % \midrule
% & \multicolumn{5}{c}{Totalcapture} \\ % Add vertical lines around Totalcapture and top horizontal line
 \hline % Add horizontal line under Totalcapture
 Method& SIP Err(deg) & Ang Err(deg) & Pos Err(cm) & Mesh Err(cm) & Jitter ($10^{2}m/s^{3}$) \\\hline
w/o Text & 9.21(+/-4.75) & 10.30(+/-4.43) & 4.19(+/-2.23) & 4.86(+/-2.54) & 1.87(+/-1.60) \\ 
w/o UGSA & 8.67(+/-4.73) & 9.94(+/-4.37) & 4.04(+/-2.22) & 4.67(+/-2.50) & 1.70(+/-1.55) \\ 
w/o HTT & 8.35(+/-4.57) & 9.70(+/-4.29) & 3.89(+/-2.10) & 4.52(+/-2.35) & \textbf{0.44(+/-1.21)} \\ \hline
Ours & \textbf{7.92(+/-4.38)} & \textbf{9.35(+/-4.10)} & \textbf{3.70(+/-2.03)} & \textbf{4.32(+/-2.29)} & 1.74(+/-1.55) \\ \hline
\end{tabular}}
\caption{Evaluation of Ablation Models on the Totalcapture Dataset. Bold numbers indicate the best performing entries.}\label{table3}
\end{table}

\begin{figure}[t]
\centering
\includegraphics[width=\columnwidth]{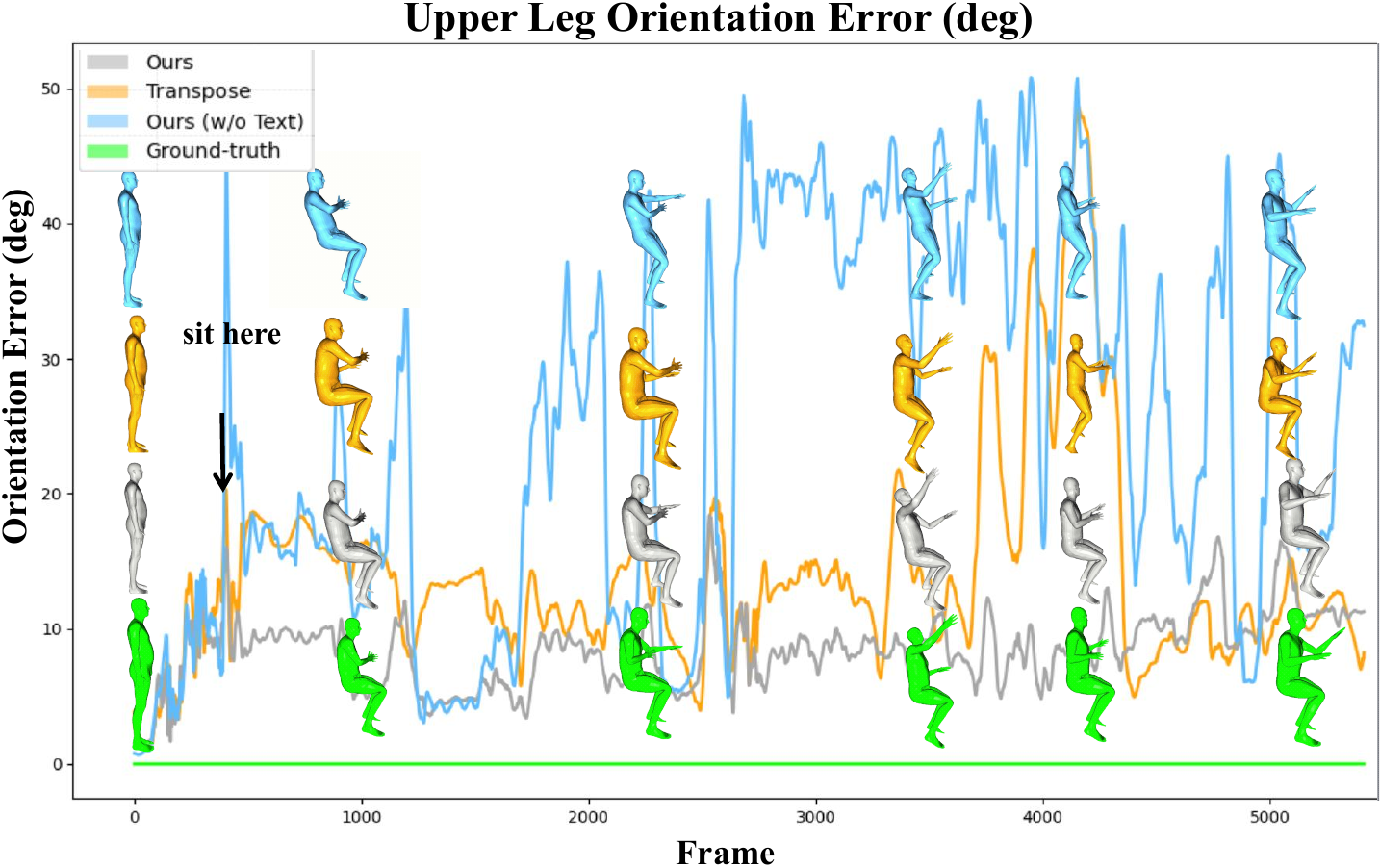} 
% Reduce the figure size so that it is slightly narrower than the column. Don't use precise values for figure width.This setup will avoid overfull boxes.(with or without text information)
\caption{We demonstrated a comparison between our method (with/without text) and Transpose in a sitting situation, focusing on the analysis of upper leg rotation error.}
\label{fig6}
\end{figure}

Without semantic information, the model's predictions fluctuate in ambiguous situations, a phenomenon illustrated in Fig. \ref{fig6} by the erratic alternation between sitting and standing positions. By incorporating a simple semantic annotation like ``sitting'', our model is able to maintain the desired sitting posture effectively.
% Fig \ref{fig5} shows a comparison between our method and the transpose technique, with and without semantic information. Without semantic information, the model's predictions fluctuate in ambiguous situations, as evidenced by the erratic alternation between sitting and standing positions. By incorporating a simple semantic annotation like 'sitting,' our model is able to maintain the desired sitting posture effectively.

% Fig \ref{fig5} presents a visual comparison between our method and the transpose technique, highlighting the differences with and without the incorporation of semantic information. As anticipated, the exclusion of semantic information from the model input resulted in fluctuating predictions for ambiguous situations - a challenge noted in previous studies. This issue led model to produce an erratic motion, continuously toggling between sitting and standing, instead of sustaining a stable sitting position. However, the incorporation of a simple semantic annotation, such as 'sitting', empowered our model to maintain the sitting posture effectively. 

\begin{figure}[t]
\centering
\includegraphics[ width=\columnwidth, keepaspectratio]{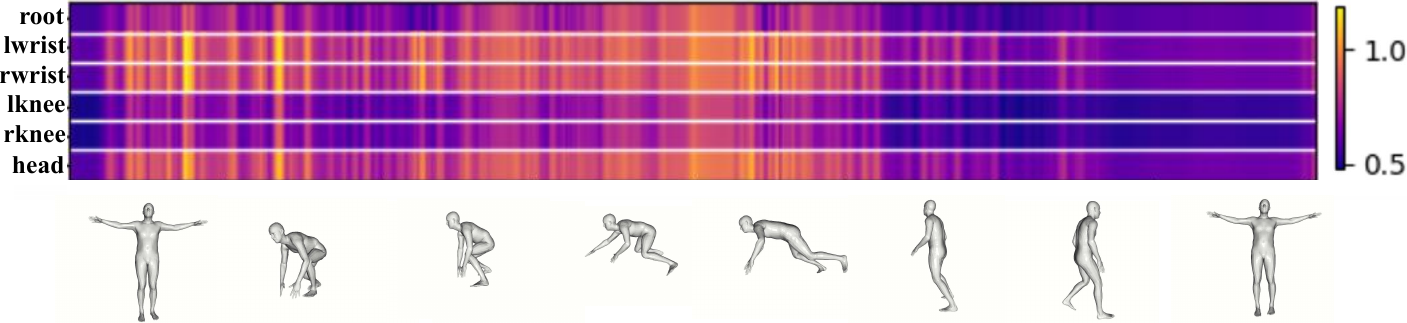} 
% Reduce the figure size so that it is slightly narrower than the column. Don't use precise values for figure width.This setup will avoid overfull boxes.
\caption{Temporal Evolution of Uncertainty Across Six Sensors: Each row represents a different sensor, with color variations indicating changes in uncertainty.}
\label{fig7}
\end{figure}

Our findings indicate that the absence of Uncertainty-guided Spatial Attention affects the accuracy of the results. Fig. \ref{fig7} illustrates how uncertainty fluctuates over time. Uncertainty increases across all sensors during complex movements like squatting and crawling, particularly in the hand regions. Conversely, a transition to a standing posture leads to a marked reduction in uncertainty, with the leg sensors showing the lowest levels.

In examining the Hierarchical Temporal Transformer (HTT), we discern that employing window attention and patch merging within this module, instead of global attention, not only curtails computational needs but also elevates performance in almost all metrics, barring jitter. We consider such a trade-off to be acceptable.
% The lack of Uncertainty-guided spatial attention notably hindered the accuracy of the outcomes. Regarding the Hierarchical Temporal Transformer, we observe that utilizing window attention and patch merge within the HTT module, as opposed to global attention, not only diminishes computational requirements but also enhances performance across all measurements, with the exception of jitter. We regard this trade-off as judicious and well-founded. 

These ablation findings affirm our approach's superior capacity for modeling sensor information and its ability to leverage semantic cues for generating more precise and natural movements.


\section{Conclusion}
In this paper, we are dedicated to addressing the ambiguity issues associated with using sparse inertial sensors for motion reconstruction.
Our approach involves enhancing the sensor data modeling capabilities and incorporating textual supervision. In the realm of sensor data modeling, we introduced an Uncertainty-guided Spatial Attention Module to model spatial relationships amongst IMUs while considering their respective uncertainty. For the modal fusion, we leverage the Hierarchical Temporal Transformer (HTT) module to achieve temporal alignment between sensor features and textual semantics. Furthermore, we employ contrastive learning to align features from both modalities in a high-dimensional space before fusion.
Experimental results have validated the effectiveness of our method. 
Looking ahead, we plan to explore the integration of real-time execution capabilities into our framework. This could include the combination of natural language reasoning with motion data, potentially utilizing ``prompt learning'' to train a decoder that performs real-time text annotation.
% The findings can be further applied to fields such as rehabilitation therapy assessment, action recognition research, and high-precision animation production.

\section{Acknowledgements}
This article is sponsored by National Key R\&D Program of China 2022ZD0118001, National Natural Science Foundation
of China under Grant 61972028, 62332017, 62303043 and U22A2022, and Guangdong Basic and Applied Basic Research Foundation
2023A1515030177, 2021A1515012285.

\bibliography{aaai24}

\end{document}